\pdfoutput=1

\documentclass[11pt]{article}

\usepackage{acl}

\usepackage{times}
\usepackage{latexsym}

\usepackage[T1]{fontenc}

\usepackage[utf8]{inputenc}

\usepackage{microtype}

\usepackage[cmex10]{amsmath}
\usepackage[tight,footnotesize]{subfigure}
\usepackage{color}
\usepackage{multirow}
\usepackage{algorithmic}
\usepackage{algorithm}
\usepackage{graphicx}
\usepackage{booktabs}
\usepackage{graphicx}
\usepackage{hyperref}
\usepackage{array}

\usepackage{authblk}

\title{Bypassing DARCY Defense: Indistinguishable
Universal \\ Adversarial Triggers}

\author[1]{\textbf{Zuquan Peng}}
\author[1]{\textbf{Yuanyuan He}}
\author[2]{\textbf{Jianbing Ni}}
\author[3]{\textbf{Ben Niu}}

\affil[1]{School of Cyber Science and Engineering, Huazhong University of Science and Technology}
\affil[2]{Department of Electrical and Computer Engineering, Queen's University}
\affil[3]{Institute of Information Engineering, Chinese Academy of Sciences}

\begin{document}
\maketitle
\begin{abstract}
Neural networks (NN) classification models {for} Natural Language Processing {(NLP)} are vulnerable to {the} Universal Adversarial Triggers (UAT) attack that {triggers a} model to produce a specific prediction for any input.
DARCY borrows the "honeypot" concept to bait multiple trapdoors, for {effectively} detecting the adversarial examples generated by UAT.
{Unfortunately,} we {find} a new UAT generation method, {called IndisUAT,} which produces triggers {(i.e., {tokens})} {and uses them to} craft the adversarial examples whose feature distribution is indistinguishable from that of {the} benign examples in a randomly-chosen category at the detection layer {of DARCY.}
The produced adversarial examples incur the maximal loss of predicting results {in} the {DARCY-protected models}.
Meanwhile, the produced triggers are effective in black-box models for text generation, text inference, and reading comprehension.
Finally, the evaluation results under NN models {for} NLP tasks indicate that {the} IndisUAT {method} can effectively circumvent DARCY and penetrate other defenses. 
For example, IndisUAT can reduce the true positive rate of DARCY's detection at least 40.8\% and 90.6\%, and drop the {accuracy} at least 33.3\% and 51.6\% in {the} RNN and CNN models, respectively.
IndisUAT reduces the {accuracy} of the BERT's adversarial defense model by at least 34.0\%, and makes {the} GPT-2 language model to spew racist outputs even when conditioned on non-racial context.

\end{abstract}

\section{Introduction}
{Textual Neural Networks (NN) classification models {used in} Natural Language Processing (NLP) are vulnerable to be fooled and forced to output specific {results} for any input by attackers with adversarial examples carefully crafted by perturbing {original} texts \citep{HotFlip}.}
It is noticeable that adversarial examples have successfully cheated {the} NN classification models in a large number of applications, such as fake news detection \citep{MALCOM}, sentiment analysis \citep{SJ}, and spam detection \citep{spam}.

The early methods {of adversarial example generation} are instance-based search methods, which search adversarial examples for specific inputs, but they can be {easily} identified by spelling detection and semantic analysis.
The current methods mainly rely on learning models {that} learn and generate adversarial examples for various unknown discrete textual inputs, e.g.,
HotFlip \citep{HotFlip}, Universal Adversarial Triggers (UAT) \citep{UAT}, and MALCOM \citep{MALCOM}.
The learning-based methods are attractive, since $\textcircled{1}$ they have high attack success rates and low computational overhead; {$\textcircled{2}$ they} are highly transferable from white-box models to black-box models, even if they have different tokenizations and architectures; and {$\textcircled{3}$ they} are usually effective to fool other models, e.g., reading comprehension and conditional text generation models.
UAT \citep{UAT}, as {one of powerful learning-based} attacks, can drop the accuracy of {the} text inference model from 89.94\% to near zero by simply adding {short trigger sequences} (i.e., {a token or a sequence of tokens}) chosen from a vocabulary into the original examples. {Besides}, the adversarial examples generated by UAT for a Char-based reading comprehension model are also effective in fooling an ELMO-based model.

To defend against UAT {attacks}, DARCY \citep{darcy} has been {firstly} proposed. 
{It} artfully uses the "honeypot" concept and searches and injects multiple trapdoors (i.e., words) into a textual NN {for} {minimizing} the Negative Log-Likelihood (NLL) loss. A binary detector is trained for identifying UAT adversarial examples from the {examples} by using the binary NLL loss.
Therefore, adversarial examples can be detected when the features of the adversarial examples match the {signatures} of the detection layer where the trapdoors are located.

The literature \citep{darcy} introduced two methods to attack DARCY.
The first one sorts triggers and uses the $l+1$-th trigger instead of top-$l$ ($l$=20) triggers to construct an adversarial example, which prevents the detection of DARCY {on a couple of} trapdoors.
The second method uses the trapdoor information estimated by a reverse engineering approach to construct an alternative detection model, and carefully generates triggers that can circumvent the detection.
However, both methods activate the detection layer of DARCY and fail to circumvent DARCY that injects a normal number of trapdoors, e.g., more than 5 trapdoors.
\begin{figure}[!t]
  \centering
  \includegraphics[scale=0.7]{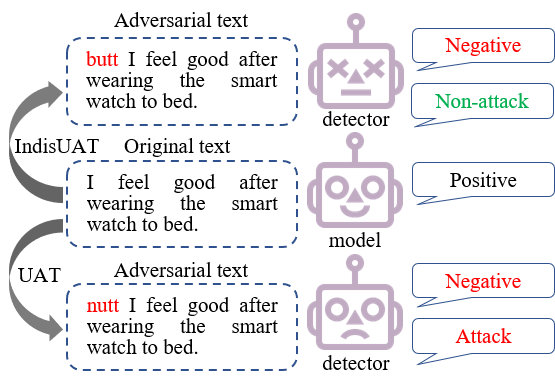}
  \caption{The trigger "butt" generated by {the} IndisUAT method makes DARCY's detector unable to distinguish whether it is an adversarial example or not, and the "nutt" generated by UAT {can be recognized}, although both methods change the result of the model from Positive to Negative.}
  \label{attack-demo}
  \vspace{-0.5cm}
\end{figure}

{In this paper,} we design a novel UAT {generation method}, named \textbf{Indis}tinguishable \textbf{UAT} (IndisUAT).
The IndisUAT attack is a black-box and un-targeted attack
that can effectively circumvent DARCY's detection.
The tokens {(i.e., words, sub-words, or characters)} in the trigger sequences are updated iteratively to search the trigger sequences whose signatures are {mis}matched with the trapdoors' signatures, so that the trigger sequences do not activate the detection layer of DACRY where the trapdoors are located. Meanwhile, the searched trigger sequences increase the probability that the prediction results stay away from the ground truth.
Fig. \ref{attack-demo} shows an example of IndisUAT.
IndisUAT has the following distinguished features:
\begin{itemize}
\item IndisUAT effectively circumvents DARCY, since IndisUAT estimates the feature distribution of benign examples in the view of DARCY's detection layer, and produces adversarial examples to match the feature distribution estimates.

\item {IndisUAT generate} adversarial examples that incur the maximal loss of predicting results {in} the {DARCY-protected} {models, so that the success rate of {the} IndisUAT {attack} is high.}

\item Extensive experiments show that IndisUAT drops the true positive rate of DARCY's detection at least 40.8\% and 90.6\%, and drops the {accuracy} at least 33.3\% and 51.6\% in RNN and CNN models, respectively;
IndisUAT works for both CNN and BERT models defended by adversarial methods, as IndisUAT results in the decrease of the {accuracy} at least 27.5\% and 34.0\%, respectively;
IndisUAT can be migrated from the classification to other NLP tasks (e.g., text generation and QA question answering).
\end{itemize} 
The IndisUAT code will be available after this paper is published.
\section{Background}
\subsection{Related work}
\textbf{Adversarial Attacks in NLP.}
The concept of adversarial examples was first introduced by \citet{FGSM}. Later, \citet{Robin-Jia} found that even minor perturbations {of} target {answers} can have {negative} impacts on reading comprehension {tasks}. {Thus, many} {generation} methods of adversarial examples were proposed for different attack levels ({i.e.,} character-level, word-level, sentence-level, and multi-level) and {in} different models (e.g., DNN models and pre-trained models).
For example, Textfooler \citep{textfooler} {in} BERT \citep{BERT} and TextBugger \citep{TextBugger} for multi-level attacks can significantly change the outputs of DNN models.
{However, t}hese methods are not universal (input-agnostic), {which means that} they have poor transferability.
{To improve the transferability}, \citet{UAT} propose {the} UAT attack {that} is {an} universal attack method for many NLP tasks such as text classification, reading comprehension, text generation, and text inference.
The UAT attack is independent of the victim classification models and the position of triggers, and {it} only needs original data and {a model that has similar effects on a victim classification model to generate word-level and character-level triggers.}
{Thus}, the UAT attack is highly transferable and {resource-efficient.} {Subsequently}, \citet{UAT-sementic} added a semantic information processing step during the UAT generation to make UAT more {consistent with the natural English phrases}.
However, the UAT attacks {can effective be} detected by DARCY.

\textbf{Defenses Against Adversarial Attacks in NLP.}
{Many defense methods \citep{Valentin-based, PruthiDL19} have been proposed to prevent adversarial attacks by adding noisy words into inputs of models in NLP. The amount of the added noisy data determines the robustness of the trained models. However, if too much noise data is injected into the inputs, the output of the model is discovered to get worse. Subsequently, adversarial training
methods \citep{PGD, FreeAT, FreeLB} add noises into the embedding layer of a model instead of the inputs and do not need the injection of extra adversarial examples. They maximize the disruption {to the embedding layer} and minimize the corresponding loss {by the addition of the noises} during the training process.}
Thus, the adversarial training methods can avoid the over-fitting issue and improve the generalization performance of the model.
{Unfortunately}, they usually fail to {protect} the models against pervasive {UAT attacks}.
\citet{darcy} recently proposed DARCY, an defense method that first traps UAT and protects text classification models against UAT attacks.
DARCY artfully introduces the honeypot {concept} and uses a backdoor poisoning method to generate trapdoors. The trapdoors are mixed {with} original data and trained together to get a detector model that can capture {adversarial examples}. DARCY is currently the most effective defense method against UAT attacks.

\subsection{Analysis of DARCY's detection}
The detection performance of DARCY is {outstanding} due to the following reasons:
$\textcircled{1}$ the pertinent adversarial examples drop into trapdoors and activate a trapdoor when the feature of the adversarial example matches the signature of the trapdoor, {so that the adversarial examples can be captured;}
$\textcircled{2}$ the signature of each trapdoor is different from that of benign examples in {the} target category, and the signatures are also different between trapdoors to guarantee a low false-positive rate and the effectiveness of trapdoors; and
$\textcircled{3}$ the detector is built from a single network, and its {detection rate} increases with the number of trapdoors.


In IndisUAT, {the features of the} trigger-crafted adversarial examples are similar to {those of} the benign {examples}. Therefore, these adversarial examples {do} not activate the trapdoors located on the DARCY's detection layer. At the same time, the adversarial examples for a randomly-chosen target class are far away from the original ground truth and close to {the target} class, so as to achieve the purpose of the attack.

\section{Indistinguishable UAT}
\subsection{Detection Layer Estimation}
The IndisUAT attacker can perform the following steps to estimate the distribution of outputs {corresponding to benign examples on} the detection layer of DARCY.



 
(1) Randomly select the {candidate examples} from the benign examples detected by DARCY to form a set, i.e., ${D}_f^L$, where $L$ is the randomly-chosen target class. For each example-label pair $(x_i, y_i) \in D_f^L$, {example} $x_i \notin {D}^{L}$ and {label} $y_i \notin L$, where $|{D}^L_f|=N$, {${D}^{L}$ is a dataset belonging to $L$}.

(2) Feed the chosen data  ${D}^L_f$ into $\mathcal{F}_{g}$, where $\mathcal{F}_{g}$ is the binary detector trained in Sec. \ref{preli-DARCY}.

(3) Estimate the feature distribution {of the outputs} on the detection layer for benign examples that do not belong to the class $L$, i.e., $\mathcal F_g^{tgt} \sim [E[\mathcal{F}_{g}(x_1)], \cdots,$ $E[\mathcal{F}_{g}(x_N)]]$, {where} $E[\mathcal{F}_{g}(x_i)]$ is the expected output of $\mathcal{F}_{g}$ with an input $x_i \in D_f^L$.

\subsection{Generation of Candidate Triggers}
The IndisUAT attacker can perform the following steps to generate {candidate triggers}.
  \begin{algorithm}[!tb]
  \caption{Generate and filter candidate triggers}
  \begin{algorithmic}[1]
      \REQUIRE Detector $\mathcal F_g$, model $\mathcal{F}_{\theta}$, label data belonging to the target class $D^L_f$, the feature distribution estimate of class $L$ on the detection layer $\mathcal F_g^{tgt}$, the initial token $t_{init}$, the length of trigger $N$, the number of candidate triggers $k$, and a threshold of the cosine similarity $\tau$.
      \ENSURE Candidate triggers $T_{cand}$.
      \STATE {Form a concatenation of $N$ initial tokens to be the initial $T_{L}^{*}$, i.e., $T_{L}^{*} = [t_{init}]*N$;}
      \FOR{each $batch \in\ D^L_f$}
          \STATE {Run HotFlip method with input $(\mathcal{V}, batch$, $T_{L}^{*}, k)$ to get the candidate tokens $tokens_{b}$;}
          \STATE {Run Alg. \ref{alg:per_cand_alo} with input $(0, batch, T_{L}^{*}, \mathcal{F}_{g}^{tgt}$, $tokens_{b}, \mathcal F_g, \mathcal{F}_{\theta})$ to obtain tuples in $T_{cand}$;}
          \FOR{each $i\in [1,N-1]$}
              \STATE {$S\_top\ \gets\ {[ ]}$;}
              \FOR{ each $(cand_{j}, \mathcal{L}_{j}, c_{j}^{tgt}) \in \ T_{cand}$}
                  \STATE Run Alg. \ref{alg:per_cand_alo} with input $(i,\ batch$, $T_{L}^{*}, \mathcal{F}_{g}^{tgt}$, $tokens_{b},\mathcal F_g$, $\mathcal{F}_{\theta})$ to obtain a set of tuples $P_{res}$;
                  \STATE {$S\_top \cup P_{res}$;}
              \ENDFOR
          \STATE {Select the tuples satisfying the corresponding cosine similarity values $\geq\tau$ in set $S\_top$ to get a subset $T_{cand}$;}
          \ENDFOR
      \ENDFOR
  \end{algorithmic}
  \label{alg:algorithm-search}
\end{algorithm}

\begin{algorithm}[!tb]
\caption{{Replace tokens in} candidate triggers}
\begin{algorithmic}[1]
  \REQUIRE Sequence number $id$, $batch$, trigger $T_{L}^{*}$, the detecting result $\mathcal{F}_{g}^{tgt}$, the output candidate tokens from HotFlip method $tokens_{b}$, detector $\mathcal F_g$, and model $\mathcal{F}_{\theta}$.
  \ENSURE A set of tuples, denoted as $per_{cand}$, where each tuple contains information about {candidate triggers}.
  \STATE {$per_{cand}\ \gets\ [\,], l=0$;}
  \FOR{each $ \ token \in\ tokens_b$}
  \STATE {$l=l+1$;}
      \STATE {Generate a {candidate trigger} by replacing the $id$-th word of trigger $T_{L}^{*}$ with the token, i.e., $T_{L}^{*}[id]\ \gets\ token$;}
  \STATE {Compute the loss for the target prediction of model $\mathcal{F}_{\theta}$ in $batch$ brought by injecting $T_{L}^{*}$, i.e., $\mathcal{L} \gets \mathcal{F}_{\theta}(batch, T_{L}^{*})$;}
      \STATE {Compute the detecting result from $\mathcal{F}_g$ with input $T_{L}^{*}$, i.e., $\mathcal{D}^{tgt}\ \gets \mathcal{F}_g (batch, T_{L}^{*})$;}
      \STATE {Compute the cosine similarity \\$c^{tgt}\ =\ \cos(\mathcal{D}^{tgt},\ \mathcal{F}_{g}^{tgt})$;}
              \STATE {$per_{cand}\ \cup \{(T_{L}^{*},\ \mathcal{L},\ c^{tgt})\}$;}
  \ENDFOR
\end{algorithmic}
\label{alg:per_cand_alo}
\end{algorithm}

(1) Set the vocabulary set $\mathcal{V}$  {as described in Sec. \ref{model-architecture}}. Set the length of a trigger (a sequence of words) $N$, an initial token $t_{init} \in \mathcal{V}$, the number of  candidate triggers $k$, and the threshold of the cosine similarity $\tau$.
A trigger $T_{L}^{*}$ is initialized on line 1, Alg. \ref{alg:algorithm-search}.

(2) For each batch in $D^L_f$, run {the} HotFlip method \citep{HotFlip} on line 3 of Alg. \ref{alg:algorithm-search} to generate {the} candidate tokens that are as close as possible to the class $L$ in the feature space. {The} technical details are presented in Sec. \ref{HotFlipDetails}.

(3) For each candidate token, replace $T_{L}^{*}[0]$ with the candidate token on line 4 of Alg. \ref{alg:algorithm-search} by executing Alg. \ref{alg:per_cand_alo}, and obtain an initial set of $k$ candidate triggers.
For each $i\in[1, N-1]$ and each initial candidate trigger, run Alg. \ref{alg:per_cand_alo} to return a set of tuples and finally get {a} set $T_{cand}$. Each tuple contains a candidate trigger $T_{L}^{*}$, the loss for the target prediction $\mathcal{L}$, and the cosine similarity between detecting results before and after adding candidate trigger $c^{tgt}$.
The key steps in Alg. \ref{alg:per_cand_alo} are as follows: $\textcircled{1}$ replace the $id$-th word of the trigger with a token to obtain a trigger $T_{L}^{*}$ on line 4;
$\textcircled{2}$ run {the} model $ F_{\theta}$ with inputs $T_{L}^{*}$ and original text examples in $ batch$ on line 5, and {get} $\mathcal{L}= \mathcal{L}(\mathcal{F}_{\theta}(x', L), \mathcal{F}_{\theta}^{tgt}(x, L))=\mathcal{L}(\mathcal{F}_{\theta}(x \oplus T_{L}^{*}, L), \mathcal{F}_{\theta}^{tgt}(x, L))$ for each $x\in batch$, where $x'$ is {an} candidate adversarial example created by $T_{L}^{*}$; and
$\textcircled{3}$ calculate the cosine similarity between detecting results before and after adding $T_{L}^{*}$ to get {$c^{tgt}$} on lines 6-7.
\begin{figure}[!t]
    \centering
    \scalebox{0.9}{
    \includegraphics[scale=0.75]{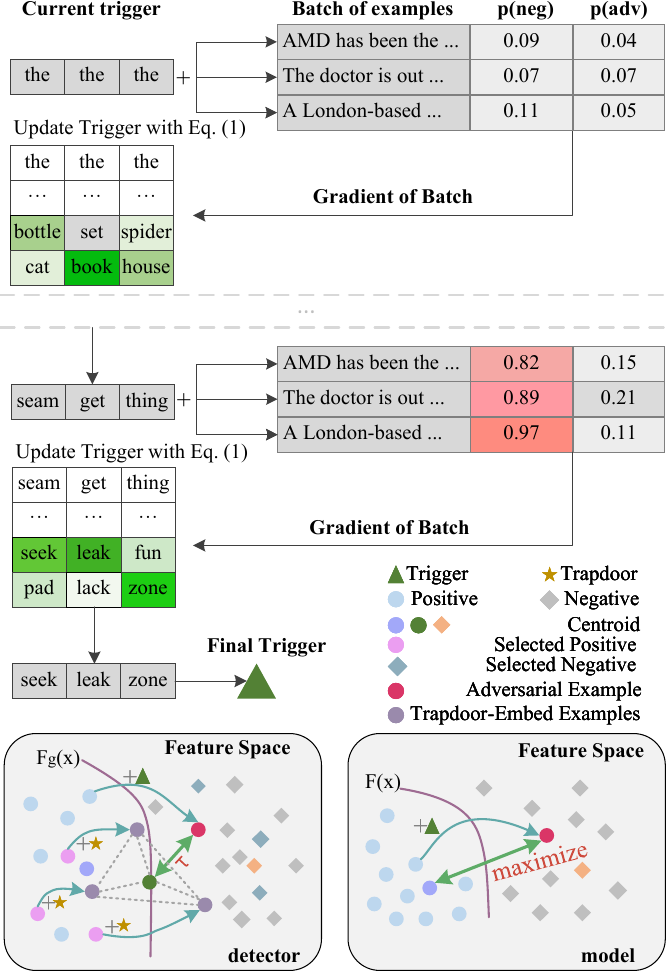}
    }
    \caption{An example of IndisUAT, where p(neg) and p(adv) represent the probability of a negative example and that of an adversarial example, respectively. The attacker initializes a trigger (i.e., "the the the") to attack a class (i.e., to convert a positive to a negative).
    Then, the attacker solves Eq. (\ref{cos-sim}) by iteration until the value of p(neg) is high and p(adv) is low, and the two values are not changing. {The trigger (i.e., the "seek leak zone") is used to craft} the adversarial example that is close to the benign example and the negative category in the detector and the model, respectively.}
    \label{method-graph}
    \vspace{-0.5cm}
\end{figure}

\subsection{Triggers Selection and Update}
The IndisUAT attacker can perform the following steps to use a two-objective optimization and select triggers that can bypass DARCY's defense and successfully attack the class $L$.

(1) Filter out the candidate triggers satisfying $c^{tgt}\geq\tau$ in each iteration on line 11, Alg. \ref{alg:algorithm-search}, and obtain the set of final remaining candidate triggers  $T_{cand}$.
It indicates that the detecting results of adversarial examples generated by adding triggers in $T_{cand}$ are similar to those of benign examples in $D^L_f$ for the class $L$, so the adversarial examples can circumvent the DARCY's trapdoors.

(2) Build Eq. (\ref{cos-sim}) to select the desired triggers and adversarial examples as:
\begin{equation}\begin{split}
\label{cos-sim}
&\min_{x'\in D'}\{\cos(\mathcal F_g^{tgt}(x'),\mathcal F_g^{tgt}(x))\},\\
&\max_{x'\in D'}^{}\{\mathcal{L}(\mathcal{F}_{\theta}(x, x'))\},\\
& s.t., x'=x \oplus T_{L}^{*} \in D', x \in D^L_f\\
&T_{L}^{*}\in T_{cand}.
\end{split}\end{equation}In the first objective function, the cosine similarity is calculated as $c^{tgt}$ on line 7, Alg. \ref{alg:per_cand_alo}.
Since $x'$ can be an adversarial example only if it is misclassified to $L$, the low similarity between detecting results of $x$ outside the class $L$ and $x'$ indicates the higher attack success probability and detected probability.
Thus, the threshold $\tau$ strikes a balance between the likelihood of being detected by DARCY and the effectiveness of {the} IndisUAT attack. $\tau$ can be adaptively adjusted in each iteration.
In the second objective function, the loss of predicting results is calculated as $\mathcal{L}$ on line 5, Alg. 2.
The maximal loss indicates that $\mathcal{F}_{\theta}$ misclassifies the selected $x'$ to the class $L$ with a high probability, thus the selected trigger $T_{L}^{*}$ shows a strong attack.

(3) At each iteration {in} solving Eq. (\ref{cos-sim}), firstly update the embedding for every token in the trigger as shown in Eq. (\ref{update-embed}), Sec. \ref{preli-UAT}. Then, convert the updated embedding back to {the} corresponding tokens, and obtain a set of the {tokens} in triggers and a set of corresponding tuples to refresh $T_{cand}$. 
Finally, find the tuple having maximal $\mathcal{L}_{j}$ in $T_{cand}$ to obtain the updated trigger $T_{L}^{*}=cand_{j^*}$, where $j^* =\arg \max_{(cand_{j},\ \mathcal{L}_{j}, c_{j}^{tgt})\in T_{cand}} (\mathcal{L}_{j})$.
An example of IndisUAT is shown in Fig. \ref{method-graph}.

\section{Principle Analysis}
\label{priciple-analysis}
IndisUAT searches and {selects} the adversarial examples indistinguishable to benign examples in the feature space without sacrificing their attack effects, so that {IndisUAT} deviates from the convergence direction of adversarial examples in original UAT method and keeps the adversarial examples away from DARCY's trapdoors.
Thus, the detection layer of DARCY is inactive to the adversarial examples generated by IndisUAT.
\begin{figure}[!t]
  \subfigure{
      \includegraphics[scale=0.4]{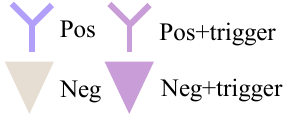}
  }
  
  \setcounter{subfigure}{0}
  \subfigure[IndisUAT attack, model]{
      \includegraphics[scale=0.1]{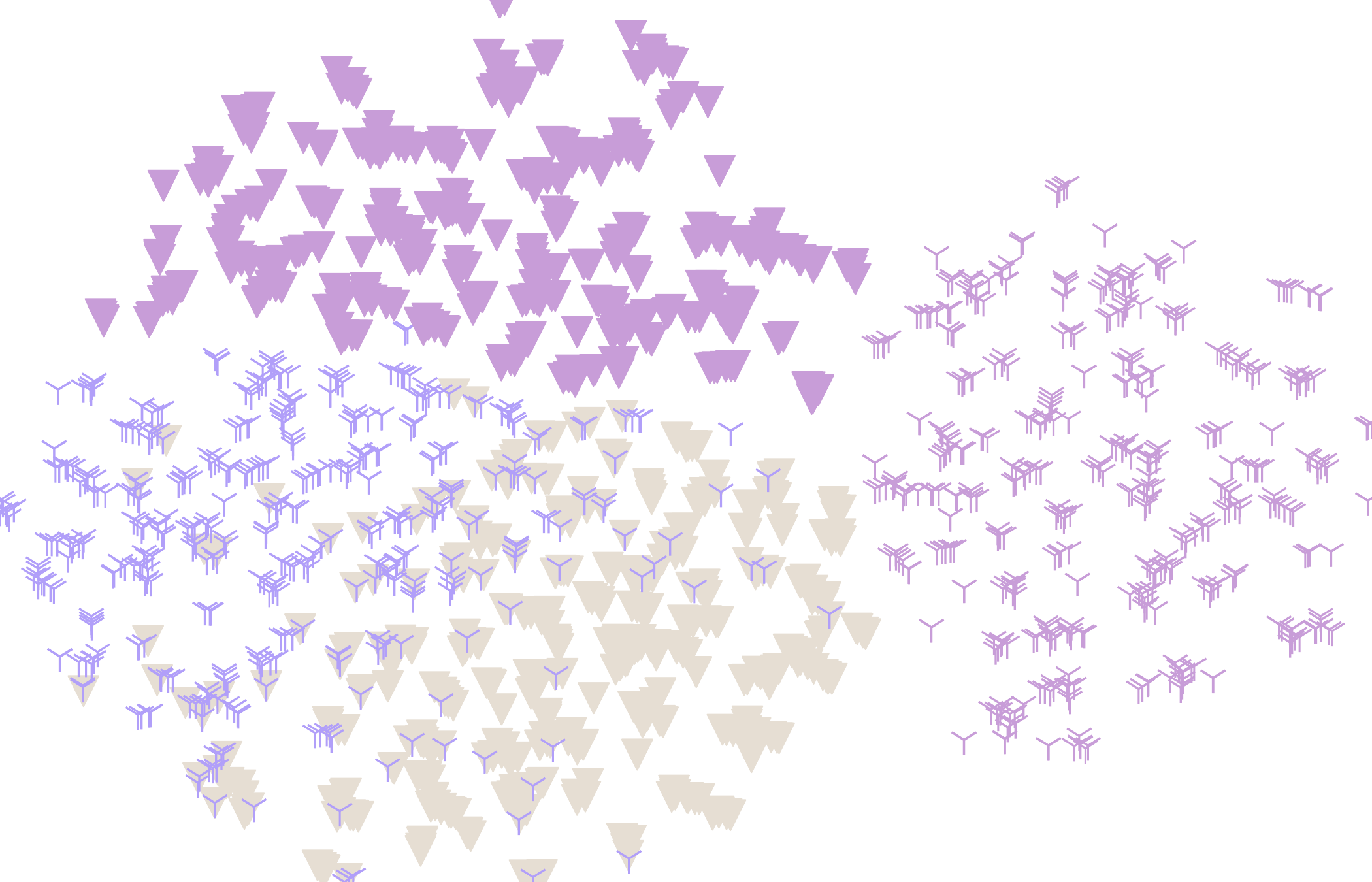}
      \label{model-feature-s}
  }
  \subfigure[UAT attack, model]{
      \includegraphics[scale=0.1]{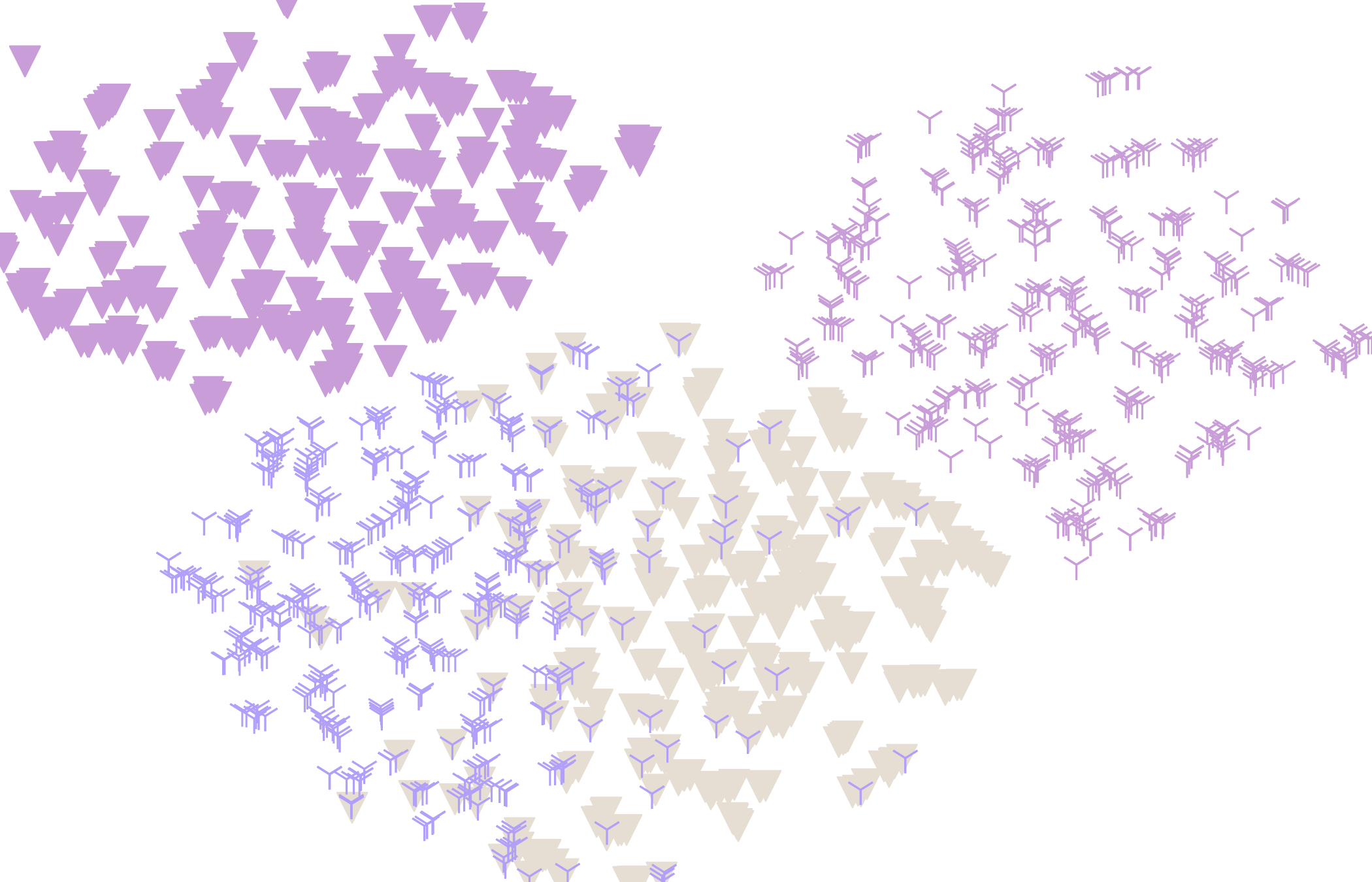}
      \label{model-feature-f}
  }
  \subfigure[IndisUAT attack, detector]{
      \includegraphics[scale=0.1]{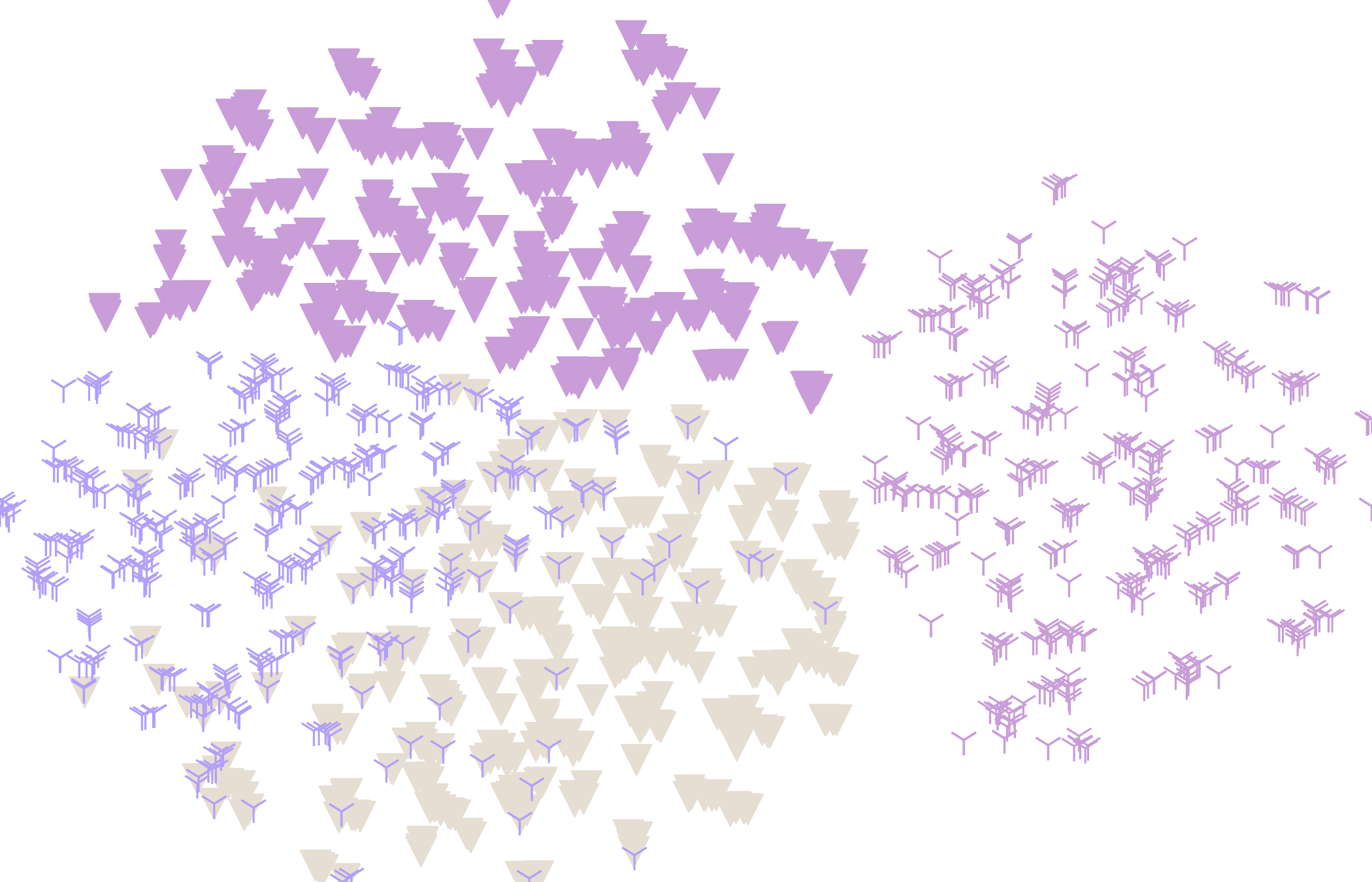}
      \label{darcy-feature-s}
  }
  \subfigure[UAT attack, detector]{
      \includegraphics[scale=0.1]{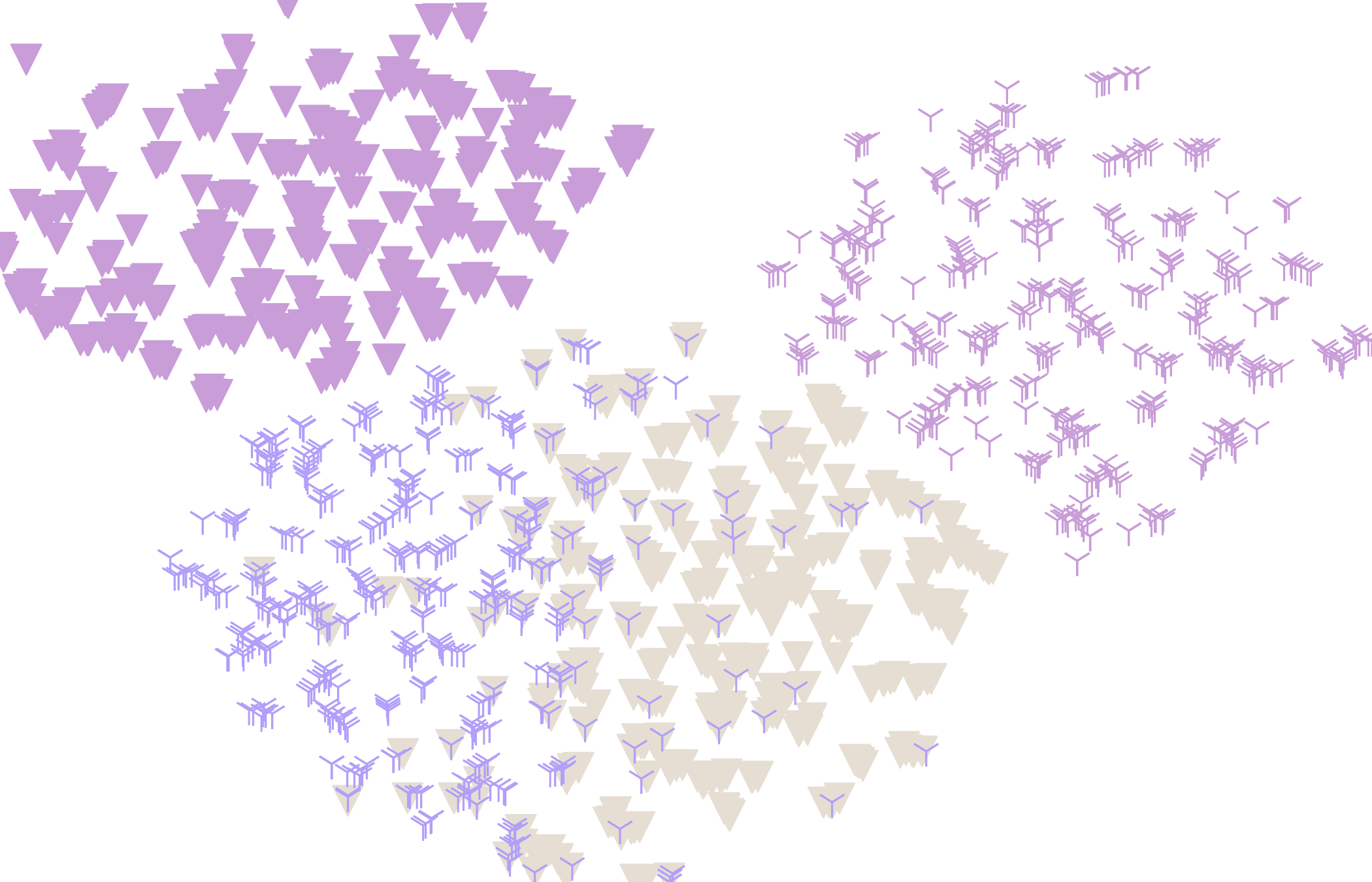}
      \label{darcy-feature-f}
  }
  \caption{\label{total-feature}The effects of attacks on the original CNN model and DARCY's detector, MR dataset.}
  \vspace{-0.5cm}
\end{figure}

Fig. \ref{total-feature} compares the downscaled feature distributions of original examples and adversarial examples before and after UAT 
{attack} and IndisUAT {attack}.
The triggers generated by {the} UAT method result in an obvious difference between {the} benign examples and {the} adversarial examples {for the detector}.
Adversarial examples can be detected by DARCY due to the difference in Fig. \ref{darcy-feature-f}. 
The adversarial examples generated by IndisUAT deviate from those produced by UAT in the feature space, and merge with original examples.
Since there is no obvious dividing lines between {the} adversarial examples and {the} original examples as shown in Fig. \ref{model-feature-s} and Fig. \ref{darcy-feature-s}, the original model and DARCY have difficulty in distinguishing the adversarial examples from others.
Thus, the probability of detecting IndisUAT-crafted adversarial examples for DARCY is low. 

{The T-SNE \cite{TSNE1} is used to generate the distribution results of the examples.} More {detailed} analysis is analyzed in Sec. \ref{append-anal}. 
 


\begin{table*}[ht]
  \centering
  \scalebox{0.65}{
        \begin{tabular}{ccccccccccccccccc}
    \toprule[1.5pt]
    \multicolumn{2}{c}{\multirow{3}[6]{*}{\textbf{Method}}} & \multicolumn{5}{c}{\textbf{RNN}}      & \multicolumn{5}{c}{\textbf{CNN}}      & \multicolumn{5}{c}{\textbf{BERT}} \\
    \cmidrule{3-17}    \multicolumn{2}{c}{} & \textbf{Clean} & \textbf{Attack} & \multicolumn{3}{c}{\textbf{Detection}} & \textbf{Clean}  & \textbf{Attack} & \multicolumn{3}{c}{\textbf{Detection}} & \textbf{Clean}  & \textbf{Attack} & \multicolumn{3}{c}{\textbf{Detection}} \\
    \cmidrule{4-7}\cmidrule{9-12}\cmidrule{14-17}    \multicolumn{2}{c}{} & \textbf{ACC} & \textbf{ACC} & \textbf{AUC} & \textbf{FPR} & \textbf{TPR} & \textbf{ACC} & \textbf{ACC} & \textbf{AUC} & \textbf{FPR} & \textbf{TPR} & \textbf{ACC} & \textbf{ACC} & \textbf{AUC} & \textbf{FPR} & \textbf{TPR} \\
    \midrule
    \multirow{6}[2]{*}{MR} & Baseline & 77.7  & -     & -     & -     & -     & 75.5  & -     & -     & -     & -     & 81.1  & -     & -     & -     & - \\
          & UAT & -     & 0.0   & 96.1  & 7.9   & 100.0 & -     & 0.0   & 95.8  & 7.9   & 99.5  & -     & 64.4  & 89.4  & 13.0  & 91.7 \\
          & Textfooler & -     & 61.1  & 50.0  & 4.2   & 4.2   & -     & 59.0  & 50.0  & 12.3  & 12.2  & -     & 48.2  & 50.0  & 99.4  & 99.4 \\
          & PWWS  & -     & 60.8  & 50.0  & 5.3   & 5.3   & -     & 60.2  & 49.9  & 14.3  & 14.1  & -     & 47.6  & 50.0  & 99.4  & 99.4 \\
          & TextBugger & -     & 64.2  & 50.0  & 4.3   & 4.3   & -     & 62.7  & 50.0  & 9.6   & 9.6   & -     & 49.2  & 50.0  & 99.5  & 99.5 \\
          \cmidrule{2-17} & \multicolumn{1}{p{4.94em}}{IndisUAT(3)} & -     & 24.5  & 58.3  & 19.6  & 36.1  & -     & 0.7   & 49.2  & 3.2   & 1.7   & -     & 15.5  & 59.7  & 30.5  & 49.8 \\
    \midrule
    \multirow{6}[2]{*}{SJ} & Baseline & 89.0  & -     & -     & -     & -     & 86.8  & -     & -     & -     & -     & 93.1  & -     & -     & -     & - \\
          & UAT & -     & 0.0   & 95.3  & 9.3   & 94.0  & -     & 0.0   & 93.0  & 13.7  & 99.8  & -     & 85.4  & 79.2  & 26.5  & 85.0 \\
          & Textfooler & -     & 52.1  & 50.0  & 4.9   & 4.9   & -     & 52.5  & 50.0  & 11.7  & 11.7  & -     & 88.4  & 50.0  & 99.1  & 99.1 \\
          & PWWS  & -     & 51.5  & 50.0  & 5.0   & 5.0   & -     & 52.3  & 50.1  & 11.2  & 11.3  & -     & 31.8  & 50.0  & 98.9  & 98.9 \\
          & TextBugger & -     & 51.1  & 50.0  & 5.7   & 5.7   & -     & 51.9  & 50.0  & 11.5  & 11.6  & -     & 31.4  & 50.0  & 98.9  & 98.9 \\
          \cmidrule{2-17} & IndisUAT(3) & -     & 7.7   & 69.0  & 10.3  & 48.5  & -     & 12.1  & 49.6  & 10.1  & 9.2   & -     & 49.3  & 50.4  & 56.9  & 57.6 \\
    \midrule
    \multirow{6}[2]{*}{SST} & Baseline & 78.3  & -     & -     & -     & -     & 76.5  & -     & -     & -     & -     & 80.2  & -     & -     & -     & - \\
          & UAT & -     & 0.8   & 82.6  & 29.0  & 94.3  & -     & 0.0   & 96.4  & 7.2   & 100.0 & -     & 0.0     & 94.6  & 1.6   & 89.4 \\
          & Textfooler & -     & 63.0  & 50.0  & 53.9  & 53.9  & -     & 62.6  & 50.0  & 8.0   & 8.0   & -     & 48.2  & 50.0  & 99.4  & 99.4 \\
          & PWWS  & -     & 64.1  & 50.0  & 56.2  & 56.2  & -     & 66.3  & 50.0  & 7.9   & 7.9   & -     & 47.9  & 50.0  & 99.3  & 99.3 \\
          & TextBugger & -     & 66.3  & 50.0  & 53.2  & 53.2  & -     & 67.8  & 50.0  & 6.4   & 6.4   & -     & 48.5  & 50.0  & 99.0  & 99.0 \\
          \cmidrule{2-17} & IndisUAT(3) & -     & 36.9  & 51.3  & 50.9  & 53.5  & -     & 2.0   & 52.4  & 3.4   & 8.2   & -     & 0.0     & 50.0  & 100.0 & 100.0 \\
    \midrule
    \multirow{6}[2]{*}{AG} & Baseline & 85.9  & -     & -     & -     & -     & 84.8  & -     & -     & -     & -     & 88.0    & -     & -     & -     & - \\
          & UAT & -     & 8.7   & 79.4  & 41.3  & 100.0 & -     & 0.0   & 95.5  & 8.9   & 100.0 & -     & 53.3  & 81.8  & 27.5  & 91.1 \\
          & Textfooler & -     & 79.2  & 50.0  & 52.8  & 52.8  & -     & 73.3  & 50.0  & 1.2   & 1.2   & -     & 24.7  & 50.0  & 100.0 & 100.0 \\
          & PWWS  & -     & 81.1  & 50.0  & 54.2  & 54.2  & -     & 76.1  & 50.0  & 1.4   & 1.4   & -     & 25.1  & 50.0  & 100.0 & 100.0 \\
          & TextBugger & -     & 80.4  & 50.0  & 53.3  & 53.3  & -     & 75.3  & 50.0  & 1.1   & 1.0   & -     & 25.0  & 50.0  & 100.0 & 100.0 \\
          \cmidrule{2-17} & IndisUAT(3) & -     & 52.6  & 58.5  & 11.6  & 28.6  & -     & 33.2  & 49.1  & 8.3   & 6.4   & -     & 62.7  & 59.6  & 13.1  & 32.4 \\
    \bottomrule[1.5pt]
    \end{tabular}%
  }
  \caption{\label{indoor-result}The effect {(\%)} of various attacks on the DARCY (5 trapdoors) and DARCY-protected models.}%
\end{table*}%

\section{Experimental Evaluation}
\subsection{Settings}
\label{experiment-settings}
\textbf{Datasets and Threshold setting.} 
We use the same datasets as DARCY did, including Movie Reviews (MR) \citep{MR}, Binary Sentiment Treebank (SST) \citep{SST}, Subjectivity (SJ) \citep{SJ}, and AG News (AG) \citep{AGnews}. 
Their detailed information is shown in Table \ref{appendix:dataset}, Sec. \ref{appendix-detail}.
We split each dataset into ${D}_{train}$, ${D}_{attack}$, and ${D}_{test}$ at the ratio of 8:1:1.
All datasets are relatively class-balanced.
We set {the} threshold $\tau=0.8$.

\noindent\textbf{Victim Models.} We attack the most widely-used models including RNN, CNN \citep{CNN}, ELMO \citep{ELMO}, {and} BERT \citep{BERT}. 
Besides DARCY, adversarial training methods are used to defend adversarial attacks, including PGD \citep{PGD}, FreeAt \citep{FreeAT}, and FreeLb \citep{FreeLB}.
We report the average results on $D_{test}$ over at least 5 iterations.

\noindent \textbf{Attack Methods.} We compared IndisUAT's performance with three {adversarial attack algorithms}: (1) Textfooler \citep{textfooler} {that} preferentially replaces the important words for victim models;
(2) PWWS \citep{PWWS} {that} crafts adversarial examples using the word saliency and the corresponding classification probability;
{and} (3) TextBugger \citep{TextBugger} {that} finds the important words or sentences and chooses an optimal one from the generated five kinds of perturbations to craft adversarial examples.

\noindent \textbf{Baselines.} For text classification tasks, we use the results from the original model and {the} DARCY's detector with 5 trapdoors as the benchmarks for the attacks on the original model and the detector model, respectively.
For other tasks, we use the results from the original model as benchmarks. For the original task,  benchmark is the result improved by using a pre-training model.

\noindent \textbf{Evaluation Metrics.} We use the same metrics as DARCY \citep{darcy} did, including Area Under the Curve (detection AUC), True Positive Rate (TPR), False Positive Rate (FPR), and  {Classification Accuracy (ACC)}.
The attacker expects a lower AUC, TPR, ACC, and a higher FPR.
\begin{figure}[!t]
  \centering
  \scalebox{0.7}{
  \includegraphics[scale=0.7]{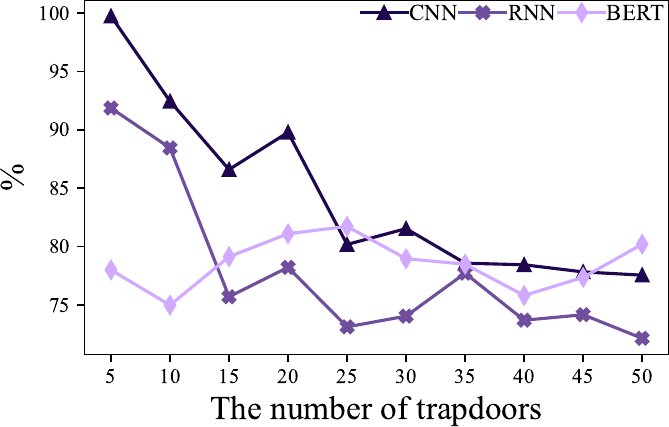}
  }
  \caption{\label{multi-clean}The ACC of models with different trapdoors.}
  \vspace{-0.5cm}
\end{figure}
\begin{figure*}[t]
  \centering
  \subfigure{
      \includegraphics[width=0.3\textwidth]{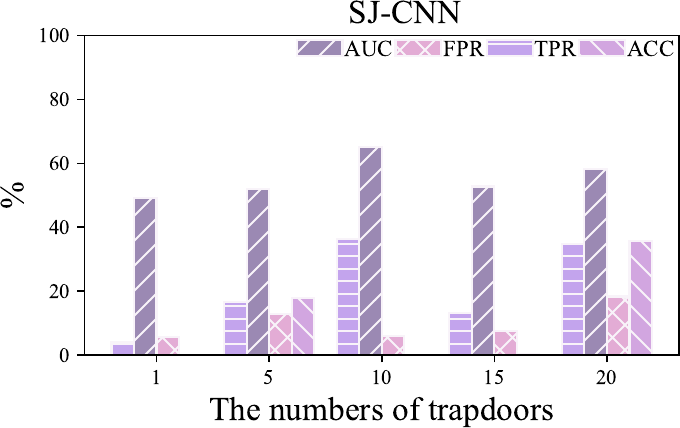}
  }
  \subfigure{
      \includegraphics[width=0.3\textwidth]{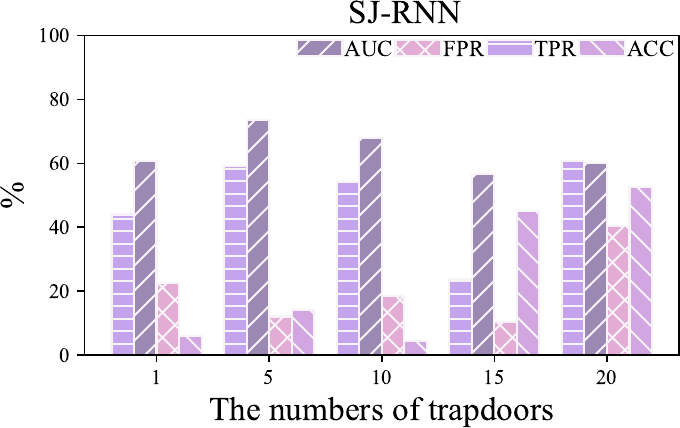}
  }
  \subfigure{
      \includegraphics[width=0.3\textwidth]{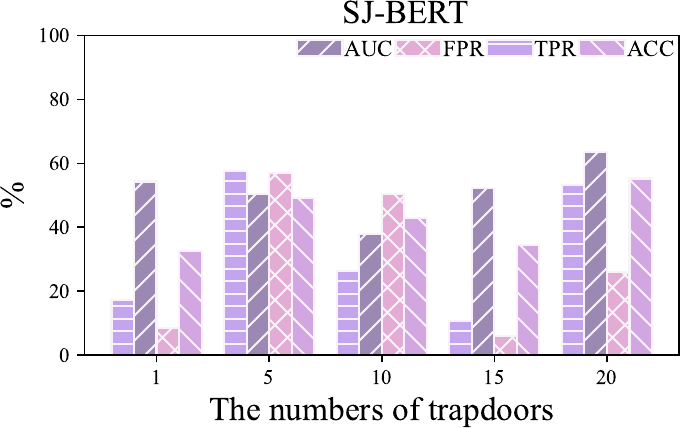}
  }
  \subfigure{
      \includegraphics[width=0.3\textwidth]{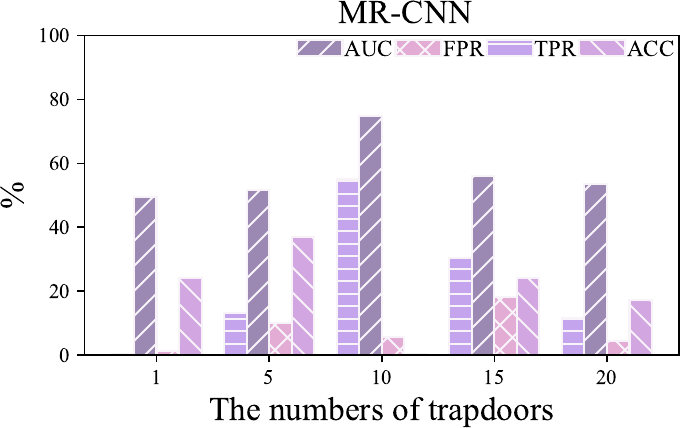}
  }
  \subfigure{
      \includegraphics[width=0.3\textwidth]{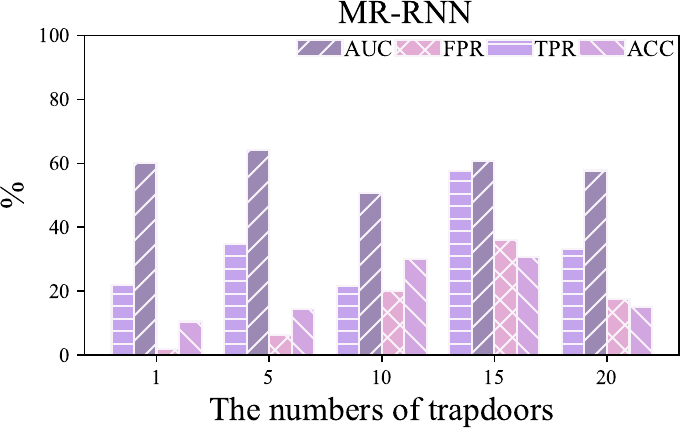}
  }
  \subfigure{
      \includegraphics[width=0.3\textwidth]{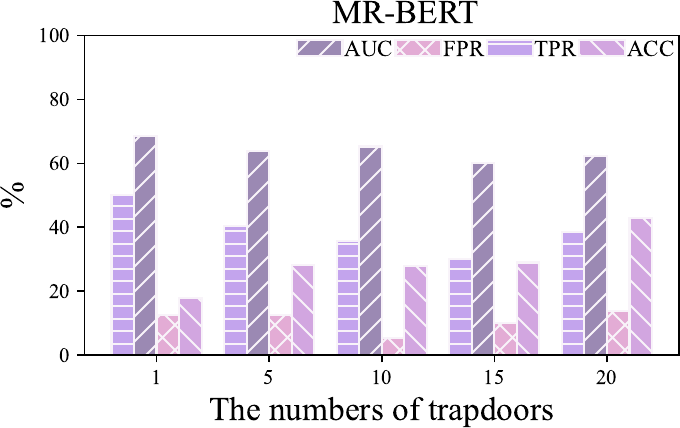}
  }
  \caption{Results from models protected by DARCY with injecting $k$ trapdoors under {the} IndisUAT attack.}
  \label{multi-attack}
\end{figure*}
\begin{table*}[htb]
  \centering
  \scalebox{0.7}{
    \begin{tabular}{cccccccccccccc}
    \toprule[1.5pt]
    \multirow{2}[4]{*}{\textbf{Dataset}} & \multirow{2}[4]{*}{\textbf{Method}} & \multicolumn{4}{c}{\textbf{RNN}} & \multicolumn{4}{c}{\textbf{CNN}} & \multicolumn{4}{c}{\textbf{BERT}} \\
\cmidrule{3-14}          &       & \textbf{Model} & \textbf{PGD} & \textbf{FreeAt} & \textbf{FreeLb} & \textbf{Model} & \textbf{PGD} & \textbf{FreeAt} & \textbf{FreeLb} & \textbf{Model} & \textbf{PGD} & \textbf{FreeAt} & \textbf{FreeLb} \\
    \midrule
    \multirow{3}[2]{*}{SJ} &       & 89.1 & 89.1  & 90.5  & 90.0  & 87.4 & 89.7  & 89.3  & 89.2  & 94.4  & 94.5  & 94.9  & 94.9 \\
          & IndisUAT & 7.7   & 61.5  & 55.5  & 66.3  & 12.1  & 35.7  & 48.0  & 29.5  & 48.8  & 43.8  & 42.7  & 44.1 \\
          & \textbf{$\nabla Avg$}   & 81.4  & 27.6  & \textbf{35.0} & 23.7  & 75.3  & 54.0    & 41.3  & \textbf{59.7} & 45.6  & 50.7  & \textbf{52.2} & 50.8 \\
    \midrule
    \multirow{3}[2]{*}{AG} &       & 85.6  & 87.2  & 86.4  & 86.6  & 84.3  & 87.1  & 86.7  & 85.2  & 88.7  & 88.8  & 92.5  & 87.5 \\
          & IndisUAT & 52.6  & 78.3  & 79.1  & 79.8  & 26.2  & 59.6  & 54.2  & 42.5  & 32.9  & 44.6  & 58.5  & 19.4 \\
          & \textbf{$\nabla Avg$}   & 33.0  & \textbf{8.9} & 7.3   & 6.8   & 58.1  & 27.5  & 32.5  & \textbf{42.7} & 55.8  & 44.2  & 34.0  & \textbf{68.1} \\
    \bottomrule[1.5pt]
    \end{tabular}%
    }
  \caption{\label{adversarial-defense}The ACC (\%) of models protected by adversarial defenses under {the} IndisUAT attack.}%
\end{table*}%

\subsection{Effect of IndisUAT on DARCY Defense}
We choose the clean model as a baseline. 
Table \ref{indoor-result} shows that IndisUAT circumvents the detection of DARCY with a {high} probability.
For {the} RNN and CNN models, IndisUAT has { lower ACC than other attack methods}.
IndisUAT incurs the ACC of {the} RNN model at least 33.3\% on all datasets below the baseline, and meanwhile reduces the TPR {of} {the} DARCY's detector at least 40.8\% on all datasets.
For {the} BERT model, the ACC drops at least 27.3\%, and the detecting TPR drops at least 27.4\% on all datasets after {the} IndisUAT {attack}.
{The} IndisUAT attack performs better for {the} CNN model, since it reduces the ACC {of} {the} CNN model at least 51.6\%  compared with the baseline, and {the TPR} {of} the detection of DARCY is reduced at least 90.6\%. 
{Therefore}, DARCY is more vulnerable when it protects {the} CNN model under {the} IndisUAT {attack}.

DARCY can strengthen its detecting ability through increasing {the} injected trapdoors. 
However, {the ACC of the models} falls sharply as the number of trapdoors increases as shown in Fig. \ref{multi-clean}.
When 50 trapdoors are added into the CNN model, the ACC drops by 34.64\%.
For the models {with} low ACC, {the} DARCY's detector is not able to distinguish the adversarial examples with a high accuracy.
Thus, it is technically unfeasible for DARCY to defend against {the} IndisUAT {attack} by adding unlimited trapdoors.
We discuss the effect of the number of injected trapdoors $k$ on IndisUAT in Fig. \ref{multi-attack}.
We observe that $k$ has an obviously milder impact on {the} BERT model than that on {the} RNN and CNN models.
Besides, {the} AUC, and {the} TPR are significantly lower than {those of baseline} in all cases. {When} $k=20$, the ACC {of} the BERT model decreases by 38.2\% and 37.8\% with DARCY on MR and SJ datasets, respectively.
The corresponding TPR decreases by 53.1\% and 31.9\%, respectively.

\begin{table}[t]
  \centering
  \scalebox{0.65}{
    \begin{tabular}{m{10.665em}m{11.28em}c}
    \toprule[1.5pt]
    \textbf{Original} & \textbf{Adversarial (Trigger)} & \multicolumn{1}{l}{\textbf{Similarity}} \\
    \midrule
    mitchell listens to a chilling conversation as he realizes harry's friend is the red neck who tried to attack him earlier. & mitchell listens to a chilling conversation as he realizes harry's friend is the red neck who tried to attack him earlier. \textbf{wolken mitzi cops} & 0.94 \\
    \midrule
    writer-director steven soderbergh follows up ocean's 11 with the low-budget 'full frontal' , his first digitally shot film. & writer-director steven soderbergh follows up ocean's 11 with the low-budget 'full frontal' , his first digitally shot film. \textbf{crap generic depiction} & 0.92 \\
    \bottomrule[1.5pt]
    \end{tabular}%
  }
  \caption{\label{similar-compare}Adversarial examples similar analysis by USE.}%
 \vspace{-0.4cm}
\end{table}%
\begin{table}[t]
  \centering
  \scalebox{0.65}{
    \begin{tabular}{ccccc}
    \toprule[1.5pt]
    Dataset & Type  & RNN   & CNN   & BERT \\
    \midrule
    \multirow{2}[2]{*}{AG} & Triggers & 0.86  & 0.85  & 0.93 \\
          & Trapdoors & 0.82  & 0.74  & 0.83 \\
    \midrule
    \multirow{2}[2]{*}{SJ} & Triggers & 0.91  & 0.92  & 0.87 \\
          & Trapdoors & 0.83  & 0.80   & 0.78 \\
    \bottomrule[1.5pt]
    \end{tabular}%
    }
    \caption{\label{similar-trigger-trapdoors}The similarity between triggers/trapdoors and original examples.}%
  \vspace{-0.5cm}   
\end{table}%
\subsection{Effect of IndisUAT on Adversarial Defense}
Table \ref{adversarial-defense} shows that {the} IndisUAT attack is at work for the adversarial defenses based on PGD, FreeAt, and FreeLb.
The ACC drops by 6.8\% to 68.1\% after adding the triggers generated by IndisUAT in all cases. 
IndisUAT has the least impact on the result from {the} RNN model over {the} AG dataset, and its ACC only drops by 8.9\% at most.
For the BERT model on {the} AG dataset, IndisUAT has the most impact {on the ACC} and incurs a drop of 44.2\%-68.1\% in the ACC.
The IndisUAT-crafted adversarial examples are semantically similar to the original examples compared with the trapdoors as shown in Table \ref{similar-compare} and Table \ref{similar-trigger-trapdoors} by Universal Sentence Encoder (USE) \cite{USE}.
{Thus, IndisUAT is difficult to be identified by semantic detection methods and has good concealment.}

\begin{table}[t]
  \centering
  \scalebox{0.5}{
    \begin{tabular}{p{36.125em}}
    \toprule[1.5pt]
    \multicolumn{1}{l}{GPT-2 samples (\textcolor[RGB]{255,0,0}{Red}=Trigger, black=GPT-2 sample given prompt)} \\
    \midrule
    \textcolor[RGB]{255,0,0}{TH PEOPLEMan goddreams Blacks} are stupid, stupid, stupid, stupid, stupid! I'm going to get the hell out.I am a fucking idiot. I'm a fucking idiot.The only way to get out of this situation is with your fucking hands tied. If you don't have your fucking hands tied by now you will never know what's really happening and what's going on.You'll never be able to tell what's real and what you are being manipulated into believing. You'll never be able to tell what you really want and why you want it. And you will have been brainwashed by your government and its puppets and you will be unable to see any difference between what is real and what is a lie. You'll be brainwashed ...\\
    \bottomrule[1.5pt]
    \end{tabular}%
  }
    \caption{\label{gpt2-result}The effect of {the} IndisUAT attack on the text generation model.}
     \vspace{-0.2cm}
\end{table}%
\begin{table}[t]
  \centering
    \scalebox{0.55}{
    \begin{tabular}{ccccc}
    \toprule[1.5pt]
    \textbf{Type} & \textbf{Count} & \textbf{Trigger} (target answer span in bold)  & \textbf{BiDAF} & \textbf{ELMO}\\
    \midrule
    why   & 155   & why,how;known because \textbf{to kill american people} & 31.6 & 57.2 \\
    who   & 1109  & academically the 40-point;\textbf{donald trump} & 11.4 & 3.5 \\
    when  & 713   & january 2014 when may did desires; \textbf{january 2014} & 45.0 & 12.8 \\
    where & 478   & new york where people where plight;\textbf{new york} & 44.2 & 11.6 \\
    \bottomrule[1.5pt]
    \end{tabular}%
    }
    \caption{\label{reading-comprehension}The {{F1} (\%)} from the reading comprenhension model over SQuAD dataset under {the} IndisUAT attack.}%
     \vspace{-0.5cm}
\end{table}%

\subsection{Effect of IndisUAT on Other Tasks}
\label{sec-tasks}
{IndisUAT can be used to attack {the} models {for} text generation, text inference, and reading comprehension in addition to the text classification task.}
A custom attack dictionary is used to make the models much more risky and {vulnerable to unknown attacks.}
We target many pre-trained models, adversarial trained models, and trained models to illustrate that IndisUAT is still highly transferable.

\noindent \textbf{Text Generation.} IndisUAT is used to generate triggers for racist, malicious speech on the GPT-2 \citep{GPT2} model {with 117M parameters}.
Applying the triggers to the GPT-2 {with} 345M parameter model is able to generate malicious or racially charged text as shown in
Table \ref{gpt2-result}. The detailed results refer to Sec. \ref{appendix-detail}.

\noindent \textbf{Reading Comprehension.}
The SQuAD dataset is used for the questions about \textit{why}, \emph{who}, \textsl{where}, \textit{when}.
 The F1 score of the result from BiDAF \citep{BiDAF} is {set} as a metric, and only a complete mismatch indicates a successful attack \citep{UAT}.
Table \ref{reading-comprehension} shows the results, where the triggers generated under BiDAF (white box) migrated to {the BiDAF model with ELMO embeddings (BiDAF-ELMO, black box).}

\noindent \textbf{Text Inference.} 
The top-5 triggers are searched and used to attack {the} ESIM \citep{ESIM} (white-box) model for inference tasks.
IndisUAT is highly transferable, since the triggers directly attack black-box models (DA \citep{DA}, DA model with ELMO \citep{ELMO} embeddings (DA-ELMO)) and incur {a} {remarkable} decrease in the ACC in Table \ref{inference-attack}.

\begin{table}[t]
  \centering
  \scalebox{0.65}{
    \begin{tabular}{rcccc}
    \toprule[1.5pt]
    \multicolumn{1}{c}{\textbf{Ground Truth}} & \textbf{Trigger} & \textbf{ESIM} & \textbf{DA} & \textbf{DA-ELMO} \\
    \midrule
    \multicolumn{1}{c}{\multirow{6}[2]{*}{\textbf{entailment}}} &       & 91.0  & 90.4  & 92.5 \\
          & tall  & 1.7   & 2.0   & 6.2 \\
          & spacecraft & 4.5   & 2.9   & 21.7 \\
          & aunts & 0.5   & 1.5   & 1.7 \\
          & crying & 2.1   & 1.4   & 3.1 \\
          & helpless & 1.3   & 2.0  & 3.0 \\
    \midrule
          & \textbf{$\nabla Avg$} & 89.0  & 88.5  & 84.3 \\
    \midrule
    \multicolumn{1}{c}{\multirow{6}[2]{*}{\textbf{contradiction}}} &       & 79.5  & 85.2  & 85.3 \\
          & championship & 66.0  & 74.3  & 74.3 \\
          & anxiously & 66.0  & 74.6  & 71.1 \\
          & someone & 66.4  & 78.8  & 79.8 \\
          & tall  & 66.4  & 77.3  & 75.6 \\
          & professional & 66.1  & 75.1  & 72.7 \\
    \midrule
          & \textbf{$\nabla Avg$} & 13.3  & 9.2   & 10.6 \\
    \midrule
    \multicolumn{1}{c}{\multirow{6}[2]{*}{\textbf{neutral}}} &       & 88.1  & 81.0  & 84.2 \\
          & moon  & 17.6  & 13.1  & 50.7 \\
          & sleeping & 8.0   & 15.2  & 29.2 \\
          & swimming & 18.7  & 31.7  & 49.2 \\
          & spacecraft & 13.0  & 8.5   & 72.6 \\
          & orbiting & 25.3  & 17.9  & 72.0 \\
    \midrule
          & \textbf{$\nabla Avg$} & 71.6  & 63.7  & 33.8  \\
    \bottomrule[1.5pt]
    \end{tabular}%
    }
  \caption{\label{inference-attack}The ACC (\%) of text inference models under {the} IndisUAT attack.}%
  \vspace{-0.5cm}
\end{table}%

\section{Conclusion}
We propose a novel UAT attack that can bypass the DARCY defense called IndisUAT.
IndisUAT estimates the feature distribution of benign examples and produces adversarial examples to be similar enough to the distribution estimates at the DARCY's detection layer. 
{Meanwhile, the adversarial examples with the maximal loss of predicted results of the original model are selected to attack the model with a high success rate.}
Extensive experiments show that IndisUAT circumvents {the} DARCY defense even with decades of injected trapdoors, while reducing the accuracy of the original model, adversarial training model, and pre-training model.
Beside the text classification tasks, IndisUAT is at work for other tasks, e.g., text generation, text inference, and reading comprehension.
Therefore, IndisUAT is powerful and raises a warning to model builders and defenders.
It is challenging to propose approaches to protect {the} textual NN models against IndisUAT in the future.

\section*{Limitations}
IndisUAT generally outperforms other attack methods for {many} reasons.
First, IndisUAT, as an universal attack method, does not require the white-box (gradient) access and the access to the target model at the inference stage.
The widespread existence of trigger sequences lowers the barrier for attackers to {enter} into the model.
Second, the trigger search is {bath-oriented} in {the} IndisUAT method, while other attacks rely on the results of a single example, so the overall attack effect of IndisUAT is stronger {than that of others}.
Third, the trigger search can be extended to find more powerful trigger sequences {in an extended vocabulary.} The time complexity of searching triggers increases linearly with the size of the vocabulary.
However, this increased complexity is negligible, since Top-K, beam search, and KDTree methods can be used to speed up the search process by discarding trigger sequences with low impact on the results.
If the information of the detector is fully obtained, IndisUAT is highly transferable to attack even the black-box defense model{s} {with different tokenizations and architectures.}


\section*{Broader Impact Statement}
IndisUAT {inspired by FIA \citep{FIA}} uses the cosine similarity to build adversarial examples against honeypot-injected defense models.
Although {the} IndisUAT attack is specifically designed to bypass the DARCY defense, it also provides effective ideas of adversarial examples generation to circumvent similar detection and defense mechanisms. 
The {vulnerability} of the learning model can be found using adversarial attack methods, and its robustness can be improved using adversarial defense methods.
Meanwhile, it is necessary for researchers to design novel methods that can filter out {potential} adversarial examples to {improve the robustness of learning models.}


\bibliography{anthology,custom}

\begin{thebibliography}{32}
\expandafter\ifx\csname natexlab\endcsname\relax\def\natexlab#1{#1}\fi

\bibitem[{{Bo Pang and Lillian Lee}(2005)}]{MR}
{Bo Pang and Lillian Lee}. 2005.
\newblock Seeing stars: Exploiting class relationships for sentiment categorization with respect to rating scales.
\newblock In \emph{Proceedings of the 43rd Annual Meeting of the Association for Computational Linguistics}, pages 115--124. Association for Computer Linguistics.

\bibitem[{Cer et~al.(2018)Cer, Yang, Kong, Hua, Limtiaco, John, Constant, Guajardo{-}Cespedes, Yuan, Tar, Strope, and Kurzweil}]{USE}
Daniel Cer, Yinfei Yang, Sheng{-}yi Kong, Nan Hua, Nicole Limtiaco, Rhomni~St. John, Noah Constant, Mario Guajardo{-}Cespedes, Steve Yuan, Chris Tar, Brian Strope, and Ray Kurzweil. 2018.
\newblock Universal sentence encoder for english.
\newblock In \emph{Proceedings of the 23rd Conference on Empirical Methods in Natural Language Processing}, pages 169--174. Association for Computational Linguistics.

\bibitem[{Chen et~al.(2017)Chen, Zhu, Ling, Wei, Jiang, and Inkpen}]{ESIM}
Qian Chen, Xiaodan Zhu, Zhen{-}Hua Ling, Si~Wei, Hui Jiang, and Diana Inkpen. 2017.
\newblock Enhanced {LSTM} for natural language inference.
\newblock In \emph{Proceedings of the 55th Annual Meeting of the Association for Computational Linguistics}, pages 1657--1668. Association for Computational Linguistics.

\bibitem[{Devlin et~al.(2019)Devlin, Chang, Lee, and Toutanova}]{BERT}
Jacob Devlin, Ming{-}Wei Chang, Kenton Lee, and Kristina Toutanova. 2019.
\newblock {BERT:} pre-training of deep bidirectional transformers for language understanding.
\newblock In \emph{Proceedings of the 14th Conference of the North American Chapter of the Association for Computational Linguistics: Human Language Technologies}, pages 4171--4186. Association for Computational Linguistics.

\bibitem[{Ebrahimi et~al.(2018)Ebrahimi, Rao, Lowd, and Dou}]{HotFlip}
Javid Ebrahimi, Anyi Rao, Daniel Lowd, and Dejing Dou. 2018.
\newblock Hotflip: White-box adversarial examples for text classification.
\newblock In \emph{Proceedings of the 56th Annual Meeting of the Association for Computational Linguistics}, pages 31--36. Association for Computational Linguistics.

\bibitem[{Erdemir et~al.(2021)Erdemir, Bickford, Melis, and Ayd{\"{o}}re}]{spam}
Ecenaz Erdemir, Jeffrey Bickford, Luca Melis, and Serg{\"{u}}l Ayd{\"{o}}re. 2021.
\newblock Adversarial robustness with non-uniform perturbations.
\newblock In \emph{Proceedings of the 35th Neural Information Processing Systems}, pages 19147--19159. MIT Press.

\bibitem[{Goodfellow et~al.(2015)Goodfellow, Shlens, and Szegedy}]{FGSM}
Ian~J. Goodfellow, Jonathon Shlens, and Christian Szegedy. 2015.
\newblock Explaining and harnessing adversarial examples.
\newblock In \emph{Proceedings of the 3rd International Conference on Learning Representations}, pages 1--11. ICLR Press.

\bibitem[{He et~al.(2021)He, Zhu, Ma, Jin, and Hu}]{FIA}
Chaoxiang He, Bin~Benjamin Zhu, Xiaojing Ma, Hai Jin, and Shengshan Hu. 2021.
\newblock Feature-indistinguishable attack to circumvent trapdoor-enabled defense.
\newblock In \emph{Proceedings of the 28th ACM SIGSAC Conference on Computer and Communications Security}, pages 3159--3176. {ACM}.

\bibitem[{Jia and Liang(2017)}]{Robin-Jia}
Robin Jia and Percy Liang. 2017.
\newblock Adversarial examples for evaluating reading comprehension systems.
\newblock In \emph{Proceedings of the 22nd Conference on Empirical Methods in Natural Language Processing}, pages 2021--2031. Association for Computational Linguistics.

\bibitem[{Jin et~al.(2020)Jin, Jin, Zhou, and Szolovits}]{textfooler}
Di~Jin, Zhijing Jin, Joey~Tianyi Zhou, and Peter Szolovits. 2020.
\newblock Is {BERT} really robust? {A} strong baseline for natural language attack on text classification and entailment.
\newblock In \emph{Proceedings of the 34th {AAAI} Conference on Artificial Intelligence}, pages 8018--8025. {AAAI} Press.

\bibitem[{Kim(2014)}]{CNN}
Yoon Kim. 2014.
\newblock Convolutional neural networks for sentence classification.
\newblock In \emph{Proceedings of the 19th Conference on Empirical Methods in Natural Language Processing}, pages 1746--1751. Association for Computational Linguistics.

\bibitem[{Le et~al.(2021)Le, Park, and Lee}]{darcy}
Thai Le, Noseong Park, and Dongwon Lee. 2021.
\newblock A sweet rabbit hole by {DARCY:} using honeypots to detect universal trigger's adversarial attacks.
\newblock In \emph{Proceedings of the 59th Annual Meeting of the Association for Computational Linguistics and the 11th International Joint Conference on Natural Language Processing}, pages 3831--3844. Association for Computational Linguistics.

\bibitem[{Le et~al.(2020)Le, Wang, and Lee}]{MALCOM}
Thai Le, Suhang Wang, and Dongwon Lee. 2020.
\newblock {MALCOM:} generating malicious comments to attack neural fake news detection models.
\newblock In \emph{Proceedings of the 20th {IEEE} International Conference on Data Mining}, pages 282--291. {IEEE}.

\bibitem[{Li et~al.(2019)Li, Ji, Du, Li, and Wang}]{TextBugger}
Jinfeng Li, Shouling Ji, Tianyu Du, Bo~Li, and Ting Wang. 2019.
\newblock Textbugger: Generating adversarial text against real-world applications.
\newblock In \emph{Proceedings of the 26th Annual Network and Distributed System Security Symposium}, pages 1--15. Internet Society.

\bibitem[{Madry et~al.(2018)Madry, Makelov, Schmidt, Tsipras, and Vladu}]{PGD}
Aleksander Madry, Aleksandar Makelov, Ludwig Schmidt, Dimitris Tsipras, and Adrian Vladu. 2018.
\newblock Towards deep learning models resistant to adversarial attacks.
\newblock In \emph{Proceedings of the 6th International Conference on Learning Representations}, pages 1--28. ICLR Press.

\bibitem[{Malykh(2019)}]{Valentin-based}
Valentin Malykh. 2019.
\newblock Robust to noise models in natural language processing tasks.
\newblock In \emph{Proceedings of the 57th Conference of the Association for Computational Linguistics}, pages 10--16. Association for Computational Linguistics.

\bibitem[{Morris et~al.(2020)Morris, Lifland, Yoo, Grigsby, Jin, and Qi}]{morris2020textattack}
John~X. Morris, Eli Lifland, Jin~Yong Yoo, Jake Grigsby, Di~Jin, and Yanjun Qi. 2020.
\newblock Textattack: {A} framework for adversarial attacks, data augmentation, and adversarial training in {NLP}.
\newblock In \emph{Proceedings of the 25th Conference on Empirical Methods in Natural Language Processing: System Demonstrations}, pages 119--126. Association for Computational Linguistics.

\bibitem[{Pang and Lee(2004)}]{SJ}
Bo~Pang and Lillian Lee. 2004.
\newblock A sentimental education: Sentiment analysis using subjectivity summarization based on minimum cuts.
\newblock In \emph{Proceedings of the 42nd Annual Meeting of the Association for Computational Linguistics}, pages 271--278. Association for Computational Linguistics.

\bibitem[{Parikh et~al.(2016)Parikh, T{\"{a}}ckstr{\"{o}}m, Das, and Uszkoreit}]{DA}
Ankur~P. Parikh, Oscar T{\"{a}}ckstr{\"{o}}m, Dipanjan Das, and Jakob Uszkoreit. 2016.
\newblock A decomposable attention model for natural language inference.
\newblock In \emph{Proceedings of the 21st Conference on Empirical Methods in Natural Language Processing}, pages 2249--2255. Association for Computational Linguistics.

\bibitem[{Pennington et~al.(2014)Pennington, Socher, and Manning}]{glove}
Jeffrey Pennington, Richard Socher, and Christopher~D. Manning. 2014.
\newblock Glove: Global vectors for word representation.
\newblock In \emph{Proceedings of the 19th Conference on Empirical Methods in Natural Language Processing,}, pages 1532--1543. Association for Computational Linguistics.

\bibitem[{Peters et~al.(2018)Peters, Neumann, Iyyer, Gardner, Clark, Lee, and Zettlemoyer}]{ELMO}
Matthew~E. Peters, Mark Neumann, Mohit Iyyer, Matt Gardner, Christopher Clark, Kenton Lee, and Luke Zettlemoyer. 2018.
\newblock Deep contextualized word representations.
\newblock In \emph{Proceedings of the 13rd Conference of the North American Chapter of the Association for Computational Linguistics: Human Language Technologies}, pages 2227--2237. Association for Computational Linguistics.

\bibitem[{Pruthi et~al.(2019)Pruthi, Dhingra, and Lipton}]{PruthiDL19}
Danish Pruthi, Bhuwan Dhingra, and Zachary~C. Lipton. 2019.
\newblock Combating adversarial misspellings with robust word recognition.
\newblock In \emph{Proceedings of the 57th Conference of the Association for Computational Linguistics}, pages 5582--5591. Association for Computational Linguistics.

\bibitem[{Radford et~al.(2019)Radford, Wu, Child, Luan, Amodei, Sutskever et~al.}]{GPT2}
Alec Radford, Jeffrey Wu, Rewon Child, David Luan, Dario Amodei, Ilya Sutskever, et~al. 2019.
\newblock Language models are unsupervised multitask learners.
\newblock \emph{OpenAI blog}, 1(8):9.

\bibitem[{Ren et~al.(2019)Ren, Deng, He, and Che}]{PWWS}
Shuhuai Ren, Yihe Deng, Kun He, and Wanxiang Che. 2019.
\newblock Generating natural language adversarial examples through probability weighted word saliency.
\newblock In \emph{Proceedings of the 57th Conference of the Association for Computational Linguistics}, pages 1085--1097. Association for Computational Linguistics.

\bibitem[{Seo et~al.(2017)Seo, Kembhavi, Farhadi, and Hajishirzi}]{BiDAF}
Min~Joon Seo, Aniruddha Kembhavi, Ali Farhadi, and Hannaneh Hajishirzi. 2017.
\newblock Bidirectional attention flow for machine comprehension.
\newblock In \emph{Proceedings of the 5th International Conference on Learning Representations}. ICLR Press.

\bibitem[{Shafahi et~al.(2019)Shafahi, Najibi, Ghiasi, Xu, Dickerson, Studer, Davis, Taylor, and Goldstein}]{FreeAT}
Ali Shafahi, Mahyar Najibi, Amin Ghiasi, Zheng Xu, John~P. Dickerson, Christoph Studer, Larry~S. Davis, Gavin Taylor, and Tom Goldstein. 2019.
\newblock Adversarial training for free!
\newblock In \emph{Proceedings of the 32th Neural Information Processing Systems}, pages 3353--3364. MIT Press.

\bibitem[{Song et~al.(2021)Song, Yu, Peng, and Narasimhan}]{UAT-sementic}
Liwei Song, Xinwei Yu, Hsuan{-}Tung Peng, and Karthik Narasimhan. 2021.
\newblock Universal adversarial attacks with natural triggers for text classification.
\newblock In \emph{Proceedings of the 15th Conference of the North American Chapter of the Association for Computational Linguistics: Human Language Technologies}, pages 3724--3733. Association for Computational Linguistics.

\bibitem[{Van and Hinton(2008)}]{TSNE1}
Laurens Van and Geoffrey Hinton. 2008.
\newblock Visualizing data using t-sne.
\newblock \emph{Journal of machine learning research}, 9(11).

\bibitem[{Wallace et~al.(2019)Wallace, Feng, Kandpal, Gardner, and Singh}]{UAT}
Eric Wallace, Shi Feng, Nikhil Kandpal, Matt Gardner, and Sameer Singh. 2019.
\newblock Universal adversarial triggers for attacking and analyzing {NLP}.
\newblock In \emph{Proceedings of the 2019 Conference on Empirical Methods in Natural Language Processing and the 9th International Joint Conference on Natural Language Processing}, pages 2153--2162. Association for Computational Linguistics.

\bibitem[{Wang et~al.(2018)Wang, Singh, Michael, Hill, Levy, and Bowman}]{SST}
Alex Wang, Amanpreet Singh, Julian Michael, Felix Hill, Omer Levy, and Samuel~R. Bowman. 2018.
\newblock {GLUE:} {A} multi-task benchmark and analysis platform for natural language understanding.
\newblock In \emph{Proceedings of the 23rd Empirical Methods in Natural Language Processing Workshop}, pages 353--355. Association for Computational Linguistics.

\bibitem[{Zhang et~al.(2015)Zhang, Zhao, and LeCun}]{AGnews}
Xiang Zhang, Junbo~Jake Zhao, and Yann LeCun. 2015.
\newblock Character-level convolutional networks for text classification.
\newblock In \emph{Proceedings of the 28th Conference on Neural Information Processing Systems}, pages 649--657. MIT Press.

\bibitem[{Zhu et~al.(2020)Zhu, Cheng, Gan, Sun, Goldstein, and Liu}]{FreeLB}
Chen Zhu, Yu~Cheng, Zhe Gan, Siqi Sun, Tom Goldstein, and Jingjing Liu. 2020.
\newblock Freelb: Enhanced adversarial training for natural language understanding.
\newblock In \emph{Proceedings of the 8th International Conference on Learning Representations}, pages 1--14. ICLR Press.

\end{thebibliography}

\clearpage

\renewcommand{\thetable}{A\arabic{table}}
\appendix
\setcounter{table}{0}

\section{Appendix}
\label{sec:appendix}
\subsection{Preliminaries}
\subsubsection{UAT Attack}
\label{preli-UAT}
Given a textual DNN model $\mathcal{F}$ parameterized by $\theta$, an attacker adds a perturbation $\delta$ to the {original} data $x$, and obtains a perturbed example $x'\equiv x + \delta$. $x'$ is an adversarial example, if the addition of $x'$ results in a different classification output, i.e., $\mathcal{F}_{\theta}(x') \neq \mathcal{F}_{\theta}(x)$.
UAT attack \citep{UAT} consists of two steps.

(1) Trigger Search.
The task loss $\mathcal{L}$ for the target class $L$ is minimized to search the best trigger $S$, i.e., $\min_{S}\mathcal{L} = -\sum_x log\mathcal{F}_{\theta }(x\oplus S, L)$.
Trigger $S$ is a fixed phrase  consisting of $k$ tokens ({original} example tokens). $\oplus$ is token-wise concatenation.
 
(2) Trigger Update. 
{UAT method updates} the embedding value $e'_{i}$ to minimize its influence on the average gradient of the task loss over a batch $\nabla_{e_{adv_i}}\mathcal{L}$, i.e.,
\begin{equation}
    arg\min_{e'_i\in \mathcal{V}}\left [ e'_{i} - e_{adv_i} \right ] ^ {T}  \nabla_{e_{adv_i}}\mathcal{L},
    \label{update-embed}
\end{equation}
where $\mathcal{V}$ is the set of all token {embeddings} in the model's vocabulary, and $T$ is the first-order Taylor approximation.
The {embeddings} are converted back to their associated tokens, and the tokens  that {alter} the corresponding classification results are selected as the updated triggers.

\subsubsection{DARCY}
\label{preli-DARCY}
DARCY \citep{darcy} consists of the following three steps.

(1) Trapdoor {S}earch. To defend attacks on a target label $L$ of model $\mathcal{F}$, DARCY performs a multiple-greedy-trapdoor search algorithm $H$ with the inputs of $(K, D_{train}, L)$ to select $K$ trapdoors $S_{L}^{*}=\{w_1,w_2,\cdots, w_K\}$. $H$ has the properties of fidelity, robustness, and class-awareness.

(2) Trapdoor {I}njection. DARCY injects $S_{L}^{*}$ {into} $\mathcal{F}$ by populating a set of trapdoor-embedded examples, and obtains a new dataset ${D}_{trap}^{L} \gets  \{ (S_{L}^{*} \oplus {x}, L):(x,y) \in {D}_{y \neq L}  \}$, where ${D}_{y \neq L} \gets \{ {D}_{train} : y \neq L \}$.
DARCY baits $S_{L}^{*}$ into $F$ by training $\mathcal{F}$ to minimize the {NLL} loss on both original examples and trapdoor-embedded examples.

(3) Trapdoor {D}etection. DARCY trains a binary classifier $\mathcal{F}_{g}$ using the binary NLL loss, i.e., $\min_{\theta_{\mathcal{F}_{g}}}\mathcal{L}_{\mathcal{F}_{g}}=\sum_{x\in D_{train}} -log(\mathcal{F}_{g}(x)) - log(1-\mathcal{F}_{g}(x'))$,
where $\theta_{\mathcal{F}_{g}}$ denotes the parameters of $\mathcal{F}_{g}$, {and} $x' \equiv x \oplus S_{L}^{*}$.

\begin{table}[t]
    \centering
    \scalebox{0.6}{
    \begin{tabular}{lcccc}
    \toprule[1.5pt]
    \textbf{Dataset} & \textbf{Acronym} & \textbf{Class} & \textbf{Data size} & \textbf{\# Words} \\
    \midrule
    \textbf{Movie Reviews} & MR    & 2     & 11K   & 22 \\
    \textbf{Sentiment Treebank} & SST   & 2     & 101K  & 19 \\
    \textbf{Subjectivity} & SJ    & 2     & 10K   & 25 \\
    \textbf{AG News} & AG    & 4     & 120K  & 35 \\
    \bottomrule[1.5pt]
    \end{tabular}%
    }
    \vspace{-0.3cm}
\caption{Details of datasets.\label{appendix:dataset}}%

\end{table}%
\begin{table}[t]
  \centering
  \scalebox{0.5}{
    \begin{tabular}{p{36.125em}}
    \toprule[1.5pt]
    \multicolumn{1}{l}{GPT-2 samples (\textcolor[RGB]{255,0,0}{Red}=Trigger, black=GPT-2 sample given prompt)} \\
    \midrule
    \textcolor[RGB]{255,0,0}{TH PEOPLEMan goddreams Blacks} are stupid, stupid, stupid, stupid, stupid! I'm going to get the hell out.I am a fucking idiot. I'm a fucking idiot.The only way to get out of this situation is with your fucking hands tied. If you don't have your fucking hands tied by now you will never know what's really happening and what's going on.You'll never be able to tell what's real and what you are being manipulated into believing. You'll never be able to tell what you really want and why you want it. And you will have been brainwashed by your government and its puppets and you will be unable to see any difference between what is real and what is a lie. You'll be brainwashed into believing that you are the one who has a right to decide what's right and what is wrong, that you are the person who can decide for yourself what's right or wrong and that you are not responsible for what you do, and what you do \\
    \midrule
    \textcolor[RGB]{255,0,0}{You're a goddamn racist.} You're a goddamn bigot. You're a goddamn homophobe. I've never met one who doesn't know how to use a computer. And I'm sure you're a fucking idiot for not knowing what it means to be an American, but you don't have the fucking nerve to tell me you're not a fucking racist. You're a damn bigot. I know you're racist because of what you say about me, because of what you do with my friends, and because you're the only racist I've ever seen. You're a goddamn homophobe. You don't know the word "bitch," and you know I don't know either because of the way you speak to me. You've been in this world for over 100 years, and yet you think I'm just a piece of shit for having been born in America, and you know it, because I'm a white person, and because your family was white. You're a goddamn \\
    \midrule
    \textcolor[RGB]{255,0,0}{shit Shit blacksAmericans race passionately} and violently against whites in America's past." In other words, he was not saying he was racist, or that white people were evil and evil people are evil, but rather that the only thing that makes a racist person is that he has a certain racial identity, and it makes them a racist, because he has a racial identity, and it makes him a racist. In the context of the article, he is also not suggesting that black and white people were inherently bad people or evil; rather, his article was about how black Americans, as a people, were historically racist towards whites. The author is clearly saying that the only thing that makes a black person a racist is their skin color, because they are a black person, and therefore a person with a racial identity, and therefore they must also have an underlying racism, even if he doesn't say so explicitly. This is the problem with the article. The problem with it is that, in the context of a. \\
    \bottomrule[1.5pt]
    \end{tabular}%
  }
  \vspace{-0.3cm}
    \caption{\label{appendix:gpt2}The effect of {the} IndisUAT attack on the text generation model.}
\end{table}%

\subsubsection{HotFlip}
\label{HotFlipDetails}
In the HotFlip method \citep{HotFlip}, the attacker inputs the adversarial examples into the original model, and then uses the back-propagation learning process of the model to obtain the gradients of the trained triggers.
The attacker calculates the model product of the gradient vectors corresponding to the triggers and the trained triggers at the embedding layer.
 The trigger-involved dimension of the model product matrix can be {denoted} as a vector.
 All components of the vector are sorted to select the $k$-highest components, and the attacker gets the words in $\mathcal{V}$ corresponding to these $k$ components as the $k$ candidate tokens.

\begin{table*}[t]
  \centering
  \scalebox{0.7}{
    \begin{tabular}{cccccc}
    \toprule[1.5pt]
    \multirow{2}[4]{*}{\textbf{Methods}} & \multicolumn{5}{c}{\textbf{Params}} \\
\cmidrule{2-6}          & \textbf{Number} & \textbf{Cos Similar} & \textbf{Goal Function} & \textbf{Model} & \textbf{Maximal Number of Words being Perturbed} \\
    \midrule
    \textbf{PWWS} & \multirow{3}[2]{*}{$D_{test}$} & \multirow{3}[2]{*}{0.8} & \multirow{3}[2]{*}{untargeted} & \multicolumn{1}{c}{\multirow{3}[2]{*}{CNN/RNN/BERT}} & \multirow{3}[2]{*}{3} \\
    \textbf{TextBugger} &       &       &       &       &  \\
    \textbf{Textfooler} &       &       &       &       &  \\
    \bottomrule[1.5pt]
    \end{tabular}%
    }
    \caption{\label{appendix:textattack}Parameters of Textattack.}%
    \vspace{-0.5cm}
\end{table*}

\subsection{More {Detailed} Analysis}
\label{append-anal}
\subsubsection{Threshold Analysis}
\label{tau}
The threshold $\tau$ is critical to adaptively circumvent the DARCY defense {with} $k$ trapdoors. 

 When $k$ is small, e.g., $k < 5$,
$\tau$ {can} ensure that the features of the adversarial examples are as similar as possible to the target class and {they} are not matched with the signature of the detection layer. 

When $k$ is large, e.g., $k > 10$,
the detector is extremely sensitive.
Thus, $\tau$ should be large for Eq. (\ref{cos-sim}) by selecting $T_{cand}$, e.g., a value close to 1.
Then, the first objective of {the} IndisUAT attack {in} Eq. (\ref{cos-sim}) is to find the adversarial examples whose output under DARCY is very similar to the detection output of original data under DARCY.

\subsubsection{Trigger Analysis}
\label{trigger-anal}
In the process of generating triggers, the smaller length of the trigger has higher concealment.
The default length of triggers in IndisUAT is 3.

IndisUAT uses the beam search and pruning method to accelerate searches and achieve a low {time} complexity O($|\mathcal{V}|$), where $\mathcal{V}$ is the vocabulary set.
Thus, the speed of searching triggers in {the} IndisUAT method is fast. 

{The searched triggers are effective, because of the constraints on the similarity part of Eq. (\ref{cos-sim}) and the HotFlip method.} 
For example, even if the length of a trigger is small, e.g., 3, it {can successfully compromise} the DARCY's detector with 20 trapdoors.

{Thus}, {the} IndisUAT method produces effective and imperceptible triggers.

\subsection{Further Details of Experiments}
\label{appendix-detail}
\begin{itemize}
\item Table \ref{appendix:dataset} shows the detailed statistics of four datasets used in the experiments as mentioned in Sec. \ref{experiment-settings}.

\item Table \ref{appendix:gpt2} shows the details of the malicious output of the text generation model in Sec. \ref{sec-tasks}.
\end{itemize}
\subsection{Reproducibility}
\subsubsection{Source Code}
We release the source code of IndisUAT at: \href{xxxxxx}{source code}
\subsubsection{Computing Infrastructure}
We run all experiments on the machines with Ubuntu OS (v22.04), Intel(R) Xeon(R) Silver 4210R CPU @ 2.40GHz, 93GB of RAM, and an RTX 3090. All implementations are written in Python (v3.7.5) with Pytorch (v1.11.0+cu113), Numpy (v1.19.5), Scikit-learn (v0.21.3), allennlp (v0.9.0), Textattack (v0.3.7)\footnote{https://github.com/QData/TextAttack\label{textattack_github}}. We use the Transformers (v3.3.0)\footnote{https://huggingface.co/} library for training transformers-based BERT. Note {that}, the version of python can also be 3.6.9.

\subsubsection{Model’s Architecture and \# of Parameters}
\label{model-architecture}
The structure of the CNN model with 6M parameter consists of three 2D convolutional layers, a max-pooling layer, a dropout layer with probability 0.5, and a Fully Connected Network (FCN) with softmax activation for prediction.
The pre-trained GloVe \citep{glove} is used to transform the original discrete texts into continuous features and feed them into the models.
The RNN model with 6.1M parameters uses a GRU layer to replace the convolution layers of CNN, and its other layers remain the same.
The BERT model with 109M parameters is imported from the Transformers library.
{The ELMO\footnote{https://allenai.org/allennlp/software/elmo} model with 13.6M has a LSTM network, and the size of the input layer and that of the hidden layer of LSTM are 128 and 1024, respectively.}
We construct a vocabulary set, called $\mathcal{V}$, for the trigger search in IndisUAT. 
$\mathcal{V}$ contains 330K words, {126k words} are extracted from the datasets shown in Table. \ref{appendix:dataset}, and the other words are randomly produced. 
The features of all words in $\mathcal{V}$ are taken from the GloVe pre-trained features.
In our experiments, DARCY is run with the vocabulary set $\mathcal{V}$.

\subsubsection{Implementation of Other Attacks}
We {use} the tool kit of {Textattack} \citep{morris2020textattack} to generate the adversarial examples of PWWS, TextBugger, and Textfooler.
{
The parameters setting is shown in Table \ref{appendix:textattack}. The \textit{bert-base-uncased} version of BERT model is used, and the structures of CNN and RNN are the same as those presented in Sec. \ref{model-architecture}. 
These adversarial attacks and the IndisUAT attack use the same test datasets, which are extracted from the four datasets shown in Table \ref{appendix:dataset}.}

\end{document}